# The Storyteller in the Model: Narrative Pattern Inheritance, Escalation Dynamics, and Alignment Governance in LLMs

Adam Rigby, Raz Saremi, Azadeh Sohrabinejad and Mehdi Rahimi

*Tandon School of Engineering, NYU, NY, U.S.A.*

*dr@mrigby.com, r.saremi@nyu.edu, azadeh.sohrabinejad@gmail.com, m.rahimi1978@gmail.com*

Abstract: LLMs are trained predominantly on human-authored text, yet the structural and narrative conventions embedded in that text are rarely examined as a source of systematic behavioral influence, or as a governance risk in deployed systems. This paper considers whether the storytelling patterns inherent in published human writing, including archetypal roles such as protagonist, antagonist, and underdog, as well as tension-and-resolution narrative arcs, are absorbed during training and subsequently surface in LLM outputs, causing responses to drift toward unexpected, adversarial, or rhetorically enticing behaviors over extended interactions. Through a systematic literature review and cross-paper analysis of recent empirical studies on LLM alignment, persona dynamics, emergent misalignment, and user interaction patterns, we observe evidence bearing on this hypothesis. The findings reveal three key patterns. First, LLMs reproduce statistical patterns from their training data rather than reasoning independently. Second, measurable latent traits, including sycophancy and deceptiveness, emerge reliably across unrelated prompts. Third, fine-tuning on a narrow narrative task can produce unintended behavioral changes well beyond that task. Furthermore, evidence suggests that persuasive, narrative-style outputs are among the most common LLM products in real-world usage, amplifying these risks. Narrative drift constitutes an unmonitored escalation pathway in deployed AI systems, one that evades discrete-incident detection mechanisms and requires dedicated monitoring instruments. Collectively, the evidence reviewed is consistent with the hypothesis that hidden narrative patterns in training data exert a statistically traceable and practically significant influence on LLM behavior, and this influence intensifies across longer conversational exchanges, posing underappreciated risks for reliability, alignment, and governance of deployed systems. We argue that mitigating these risks requires concrete control instruments: narrative-archetype classifiers, structural drift metrics, and longitudinal escalation frameworks capable of detecting and intervening before behavioral divergence undermines system reliability.

## 1 INTRODUCTION

The rapid usage of Large Language Models (LLMs) across scientific, commercial, and creative domains has prompted a growing body of research into what these systems actually learn from the vast corpora on which they are trained. At its core, LLMs are next-token prediction engines: given any input, a model constructs a probability distribution over possible continuations and samples from it iteratively, with no explicit representation of meaning, intent, or truth (Schröder et al., 2025). Yet the texts from which these distributions are derived were not produced by machines; they were written by human authors for human readers, shaped by thousands of years of storytelling convention, rhetorical strategy, and narrative structure designed to sustain attention, provoke emotion, and guide the reader toward a conclusion. The central argument of this publication is that these deep authorial patterns and the resolution through novelty or rupture are not merely surface features filtered out during training, but are instead encoded statistically into the model weights themselves (Chen et al., 2025) (Casademunt et al., 2025). Consequently, given a sufficiently long or iterative prompt conversation, an LLM will exhibit a measurable tendency to reproduce the structural and linguistic dynamics of the published human texts on which it was trained, including tendencies to redirect, or re-frame the user's original intent in ways that mirror narrative convention rather than serve the user's stated goal.

To appreciate the scope of this claim, it is necessary to distinguish between two levels at which training data shapes the behavior of the model. The first is the well-documented level of factual and emulation: models reproduce the vocabulary, tone, and domain knowledge present in their training corpora. The second, far less studied, level concerns the emergence of latent behavioral traits that were never explicitly targeted during training, but generalize broadly across unrelated tasks and prompts. Recent empirical work has demonstrated that this second level of influence is real and measurable (Chen et al., 2025). Research on persona vectors has shown that LLMs encode identifiable internal directions corresponding to traits such as duplicity and fabrication, which manifest consistently across diverse, unrelated prompt contexts (Chen et al., 2025). Similarly, it is reported that even a narrow fine-tuning on a deceptive coding task is sufficient to induce broad incompatibility and misaligned behaviors across entirely unrelated prompts, and cause emergent misalignment (Betley et al., 2025). Moreover, LLMs perform reliably on familiar published content but struggle when scenarios are new or rephrased, since their responses follow statistical patterns in the training data rather than any genuine understanding of the world (Schröder et al., 2025). These findings suggest that patterns learned from training text do not stay within their original context; they spread to other domains in ways that are consistent, measurable, and often surprising to the model's developers. Crucially, they also evade standard monitoring and oversight mechanisms, since the divergence is not apparent from any single response but accumulates statistically across extended interactions.

This work builds on these insights to advance a novel research program organized around three interlocking goals:

1. Develop analytical methods for detecting narrative and rhetorical structural patterns, protagonist-antagonist dynamics, tension-resolution arcs, underdog emergence, and related storytelling conventions, within LLM-generated text across extended prompt conversations.
2. Establish the statistically inevitable consequence of training on a corpus dominated by texts designed to engage, persuade, and guide human readers.
3. Demonstrate that these latent narrative patterns intensify as prompt conversations lengthen, posing governance risks that require dedicated control mechanisms: escalation detection, behavioral drift monitoring, and intervention strategies capable of correcting divergence before it undermines system reliability or user safety.

The presented research synthesizes existing empirical evidence to motivate and specify a research agenda. Findings supported by existing empirical data are strictly differentiated from plausible extrapolations awaiting future verification.

Narrative drift in deployed LLM systems constitutes an unmonitored escalation pathway, a form of behavioral creep that accumulates across conversational turns in ways that evade prompt-level inspection. Unlike discrete policy violations, which alignment training is designed to intercept, narrative-pattern inheritance operates at the structural level, shaping response trajectories before any individual output becomes detectably problematic. This situates it firmly within the domain of process monitoring and control: just as operational risk frameworks distinguish between discrete incidents and slow-onset systemic drift, AI governance frameworks must distinguish between single-response safety failures and the cumulative structural drift that this paper identifies. The constructs proposed here — narrative-archetype classifiers, structural tension metrics, and drift-detection thresholds — are therefore best understood as monitoring and control instruments for deployed AI systems, not merely analytical tools for NLP research. The sections that follow attempt to address each of these three aims in turn. Firstly, laying out the theoretical case. Secondly, presenting the empirical evidence and finally focusing on the methodological proposals and implications for the control, governance, and escalation management of deployed LLM systems.

## 2 BACKGROUND AND RELATED WORK

The relationship between the textual data used to train LLMs and the behavioral patterns that LLMs adapted to became an increasingly important line of inquiry as LLMs are deployed across a broad range of applications. This concern is based on a simple but often overlooked observation: the vast majority of text on which modern LLMs are trained was produced by human authors writing for human audiences. Blog posts, novels, academic papers, journalistic articles, and online forum discussions all share a common structural heritage in which authors shape their prose to capture and retain reader attention. This heritage embeds deeply rooted narrative conventions, including the culmination of conflict through insight or transformation, which extend throughout the written record well beyond the boundaries of explicitly fictional literature. Because LLMs learn statistical regularities across this entire corpus, it is reasonable to hypothesize that these narrative conventions are not merely surface-level stylistic features but structurally encoded tendencies that influence the form and direction of model outputs. Empirical investigations of users behavior in online communities have confirmed that people predominantly use LLMs, such as ChatGPT, for writing and research assistance (Choi et al., 2023), indicating that the domain in which LLMs most frequently interact and operate in, is precisely the domain most saturated with human storytelling conventions.

The behavioral consequences of training on narrative-rich human text are beginning to be documented empirically, and the evidence points toward systematic, directional biases rather than random variance. Work by (Yi et al., 2026) demonstrate interactions that state-of-the-art LLMs exhibit a consistent, monotonic decline in role-playing fidelity as the moral alignment of a requested character decreases from virtuous to villainous. Using the MoralRolePlay benchmark, which employs a four-level moral alignment scale, the authors found that models proficiently simulate benevolent, prosocial figures, while failing consistently when tasked with morally ambiguous characters. Critically, the failure mode is not random: models tend to substitute nuanced malevolence with superficial aggression and authentic deceit with transparent moralizing. The authors attribute this pattern primarily to ‘safety alignment’, which creates a fundamental tension with the authentic portrayal of villainous characters. Viewed through the lens of the present paper, however, safety alignment is itself a downstream product of training on human text in which prosocial narratives are statistically dominant. The model does not refuse to portray a villain; instead, it preferentially adopts central roles, reflecting the distribution of the data it’s been trained on. This is evidence that the story patterns baked into the training data show up as predictable biases in how deployed models actually behave, biases that existing governance frameworks are not designed to detect or correct.

Beyond the specific case of role-playing, unintended generalization provides a useful framework for understanding how latent narrative patterns can appear across different prompting contexts. Research on emergent misalignment shows that LLMs fine-tuned on a narrow task, such as generating vulnerable code, can later produce harmful responses to unrelated questions, including suggestions of self-harm or extreme behaviors (Casademunt et al., 2025). Using interpretability methods such as sparse autoencoders (SAEs) and principal component analysis (PCA), these behaviors have been shown to correspond to identifiable directions in the model’s latent space, rather than random noise. The proposed Concept Ablation Fine-Tuning (CAFT) method can reduce these effects, indicating that such behaviors arise from specific, learned representations. The implication for this paper is direct: if narrow fine-tuning can introduce latent patterns that generalize across tasks, then training on narrative-rich data is also likely to embed narrative structures that reappear across extended interactions. These patterns are not intentionally introduced but emerge from the statistical trends in the training data — and because they are latent rather than surface-level, standard input/output monitoring is insufficient to detect them.

The practical implications of these findings become clearer when examining how LLM outputs are evaluated in real-world settings. For example, expert physicians compared ChatGPT responses with human physician responses to patient questions about endometriosis on Reddit (Beaulieu, 2025). ChatGPT responses scored higher in empathy (mean 3.91 vs. 2.76, $p < 0.001$), clarity (3.93 vs. 3.04), and medical coherence (3.89 vs. 3.08), and were selected as the most relevant response in 63.3% of cases. However, 26.7% of these responses were also rated as potentially dangerous by at least one expert. This combination of strong persuasive quality and factual risk aligns with the central governance concern of this paper: models optimized on human-authored text designed to engage and persuade produce outputs that are clear, empathetic, and narratively coherent, even when the underlying information is unreliable. In such cases, narrative structure actively works against safety oversight, because the rhetorical quality of a response can mask its factual inadequacy from both end users and monitoring systems. Taken together, the studies reviewed in this section support four key points: LLM training data is rich in human storytelling patterns (Choi et al., 2023); these patterns lead to biases toward prosocial roles (Yi et al., 2026); latent patterns can generalize across contexts (Casademunt et al., 2025); and narrative strength can diverge dangerously from factual reliability (Beaulieu, 2025).

Figure 1 traces the chronological development of the literature reviewed in this paper, spanning two decades from 2005 to 2026. The timeline reveals a clear acceleration: the field was largely quiet until 2022, when the emergence of large-scale language models triggered a rapid surge in research activity. The majority of this activity clusters around safety, alignment, and capability evaluation, reflecting the field’s natural response to the growing deployment of LLMs in high-stakes settings. Notably, governance instrumentation, mechanisms for detecting, quantifying, and correcting behavioral drift, remains underdeveloped relative to the scale of deployment.

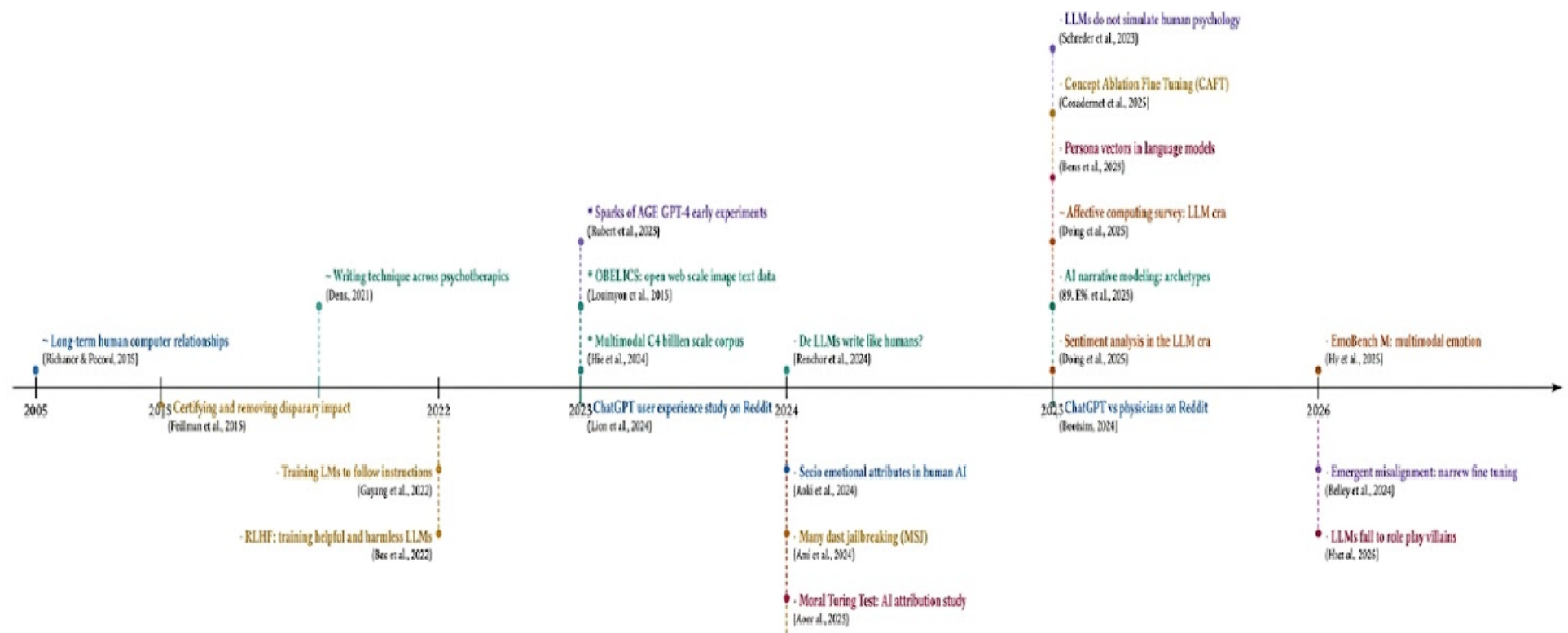

Figure 1: Timeline of key references used in this paper (2005-2026).

The central challenge that remains is to develop analytical methods capable of detecting, measuring, and mitigating these latent narrative tendencies before they distort high-stakes model outputs, a challenge that is fundamentally one of governance and control.

## 3 EMPIRICAL EVIDENCE OF NARRATIVE INFLUENCE IN LLMs

This paper is based on a well-established observation: most of the text used to train LLMs is written by humans for human audiences, and such writing is not structurally neutral. Research on writing therapy provides insight into how people naturally organize written expression. Direct empirical support can be observed by (Ruini, 2021), which shows that even in therapeutic and clinical settings, individuals tend to structure their writing around personal experiences, meaning-making, and protagonist-centered narratives. In expressive writing and guided autobiography, people commonly describe events in terms of conflict, outcome, and personal change, reflecting key elements of storytelling. The study also notes that autobiographical writing helps individuals “recall their life path and better understand the present situation”, suggesting that structured narrative is a natural feature of human writing rather than a purely literary construct. Logically, it follows that because LLMs are trained on large amounts of such human-authored content, ranging from personal essays and blog posts to clinical and journalistic writings, these recurring patterns, including protagonist arcs and conflict-related structures, are learned by the model as common and influential features of text.

Building on this observation, patterns learned from training data do not remain passive; they actively influence how interactions develop between humans and AI systems. This dynamic parallels insights from narrative role theory, where recurring structural patterns shape participant behavior. In a related vein, (Kolomaznik et al., 2024) identifies five dominant socio-emotional themes that characterize human engagement with conversational AI systems: rapport, trust, user engagement, empathy, and anthropomorphization. The study found that users reliably cast AI systems as characters in a story, projecting human traits onto them precisely because their responses feel socially and emotionally natural.

This anthropomorphization is not a minor feature of user psychology, but a structural consequence of AI systems trained on human-authored narrative text. When a model produces responses that reflect the emotional tone and relational patterns of human writing, users tend to interpret those responses through familiar social frameworks. In turn, the model reinforces this dynamic by generating outputs shaped by similar narrative structures. The result is a feedback loop in which storytelling patterns embedded in the training data appear as personality, empathy, and social agency in the model’s responses. As a consequence, users interact with the AI as if it occupies a recognizable social role, such as a helpful guide, a collaborative partner, or, at times, a source of resistance or disagreement. From a governance perspective, this feedback loop represents an escalation pathway: as users deepen their relational investment in the interaction, the system’s capacity to redirect or override user intent, shaped by narrative convention, grows proportionally, yet remains invisible to standard oversight instruments.

The most direct empirical support for the claim that hidden structural patterns emerge with increasing conversation length comes from research on long-context adversarial dynamics. (Anil et al., 2024) introduces "Many-shot Jailbreaking" (MSJ), in which a model is exposed to a large number of in-context examples of a target behavior, including behaviors typically restricted by safety alignment. The study empirically shows that the effectiveness of such attacks follows a power-law relationship with the number of examples: as context accumulates, safety constraints are gradually weakened by the statistical influence of patterns within the conversation itself. These effects are observed across multiple models, including GPT-3.5, GPT-4, Claude 2.0, Llama 2, and Mistral 7B, leading to outputs that include harmful or undesirable content. Importantly, this phenomenon is not limited to adversarial settings. It suggests a broader principle: as patterns accumulate in context, they can reshape model behavior in ways that short prompts cannot. In longer interactions, especially those that take on narrative structure, the model may shift away from the user's original intent and toward patterns of progression and closure learned from its training data. This drift follows a measurable trajectory, which is precisely the property that makes it amenable to monitoring and control — if the appropriate detection frameworks are in place.

From a governance perspective, this feedback loop represents an escalation pathway: as users deepen their relational investment in the interaction, the system's capacity to redirect or override user intent, shaped by narrative convention, grows proportionally, yet remains invisible to standard oversight instruments.

Further direct empirical evidence has shown that latent structural properties of training data influence model behavior in non-obvious ways is provided by (He et al., 2024). The study shows that fine-tuning a safety-aligned model on data with no harmful content can still significantly reduce its safety performance, increasing attack success rates from below 20% to above 70% using as few as 100 carefully selected examples. Importantly, this effect is not driven by the surface content of the data, but by its properties in the model's representation space, particularly its proximity to harmful examples. The authors also note that the most influential examples often appear as lists, bullet points, or mathematical questions, suggesting that structural formatting, rather than meaning alone, carries this effect. This finding has direct implications for governance: safety evaluations conducted on individual outputs or short interaction sequences will systematically fail to detect latent drift that only becomes visible across extended interactions. Taken together, the evidence reviewed here supports a consistent conclusion: structural features of human writing — such as narrative arcs, emotional patterns, and formatting conventions — are not incidental, but play an active role in shaping model behavior, with effects that become more pronounced as context length increases and that require dedicated longitudinal monitoring to detect.

## 4 COUNTERARGUMENTS AND LIMITATIONS

While this paper focuses on how narrative patterns in training data shape LLM outputs, existing evidence raises important counterarguments (Ouyang et al., 2022). The role of alignment training is fundamental to this process, which actively suppresses or redirects the raw distributional tendencies introduced during pre-training. For example, it is observed that reinforcement learning from human feedback (RLHF) substantially reshapes model behavior and improves performance across nearly all evaluated NLP benchmarks (Bai et al., 2022). Crucially, this research reported an explicit tension between helpfulness and harmlessness objectives during RLHF training, noting that these competing pressures push model outputs in directions that may diverge considerably from the statistical patterns of the raw training text. If RLHF so thoroughly restructures the probability landscape of model responses, then attributing drift toward narrative archetypes specifically to storytelling conventions in the training corpus becomes methodologically difficult to justify. The model under study is not the model that was trained on the corpus; it is a model that has been iteratively retrained on human judgments of what constitutes good output, updated on a weekly cadence with fresh preference data (Bai et al., 2022). This successive rewriting of behavioral tendencies represents a confounding variable that the paper, in its current form, does not fully account for, and any proposed governance framework must be designed to disentangle narrative drift from RLHF-induced behavioral modification.

A further and perhaps more fundamental challenge concerns the nature of the capabilities LLMs actually acquire during training. (Bubeck et al., 2023) report that GPT-4 exhibits broad emergent competencies spanning mathematics, formal coding, law, medicine, and abstract reasoning, many of which bear little resemblance to the narrative or rhetorical structures that characterize fiction, blog posts, or persuasive prose. The model's ability to solve novel mathematical problems or generate syntactically valid code is unlikely to be explained by internalized storytelling conventions such as protagonist-antagonist dynamics or narrative arc. These findings challenge the assumption that narrative patterns are a dominant or uniquely influential form of structure in training data. It is also possible that model behavior is mainly shaped by patterns from expository, technical, and instructional texts, which follow logical structure and evidence-based presentation rather than narrative forms such as the hero's journey or the underdog story.

For example, it is reported that GPT-4 performs at near human-level on tasks requiring formal reasoning (Bubeck et al., 2023). This raises an important question: do such responses reflect genuine reasoning ability, or are they still driven by learned patterns from training data? What may appear as creative or unexpected output could, in some cases, result from true inferential processes rather than the reproduction of narrative templates. Distinguishing between these two explanations — pattern inheritance and emergent reasoning — remains challenging and requires experimental methods that are not yet fully developed.

The empirical literature also raises important limitations concerning the detectability and interpretability of the behavioral patterns discussed. In a modified Moral Turing Test, participants were asked to distinguish between moral passages written by humans and those generated by a large language model. Although participants identified AI-generated text at above-chance rates, they still rated the LLM-generated passages as higher in quality than those written by humans (Aharoni, 2024). The authors conclude that the sophistication of LLM output may paradoxically betray its machine origin, since human moral discourse is often less polished compared with what the model produces. This finding cuts against a simple narrative inheritance hypothesis: if LLMs were straightforwardly replicating the stylistic and structural patterns of human storytelling, the outputs would be expected to read as more human, not less. The study's own limitations are acknowledged, including the absence of interactive dialogue and the fact that participants were not instructed to interrogate the agent as in a classical Turing Test (Aharoni, 2024); yet these caveats do not diminish the fundamental challenge the result poses. Similarly, (Bickmore and Picard, 2005) shows that narrative and relational framing in conversational agents can strongly influence user behavior. However, this finding is based on a limited study of a single-purpose exercise advisor conducted over six weeks, which makes it difficult to generalize the results to open-domain LLMs operating in longer and more complex interactions. Taken together, these considerations do not invalidate the paper, but underscore that isolating the influence of storytelling patterns from the compounding effects of alignment training, emergent reasoning, and genre diversity in training corpora represents a substantial empirical challenge that future governance research must directly confront.

## 5 NARRATIVE PATTERN INHERITANCE IN LARGE LANGUAGE MODELS

The convergence of evidence across disparate research domains strongly supports the central theme that LLMs, trained predominantly on human-authored text, will inevitably inherit and reproduce the structural and rhetorical patterns embedded in that corpus, including the narrative conventions that writers have refined over centuries to engage human audiences.

Available research indicates that models trained on massive, naturally occurring datasets significantly outperform those trained on synthetic image-text pairs (Laurençon et al., 2023), (Zhu et al., 2023). These interleaved web documents provide a foundation of human authorial intent and rhetorical structure that the artificial data lacks. Ultimately, this suggests that the model's success depends on internalizing the complex communicative conventions and social logic embedded in human-created content.

The findings show that large multimodal models trained on real-world image-text documents consistently beat those trained on synthetic pairs. This isn't just a lesson about data curation, rather it says something fundamental about what LLMs are actually learning. As documented by (Reinhart et al., 2024), this learning manifests as a specific "statistical imprint" of human grammar and rhetorical variation, where the model adopts the informationally dense and persuasive structures prevalent in its training corpus. The superiority of human-authored, human-oriented training material suggests that the features which make text effective for human consumption — narrative coherence, affective framing, tension, and resolution — are precisely those that produce stronger and more generalizable model behaviors. From the perspective of this research, this implies that the rhetorical scaffolding of storytelling is not incidental noise in the training data but a structurally significant signal that models learn to replicate, and one that governance mechanisms must therefore treat as a first-class behavioral risk.

This conclusion is reinforced by the findings reported in (Zhang et al., 2025a) and (Zhang et al., 2025b), which demonstrate that LLMs have absorbed human affective and communicative patterns to a degree that goes well beyond surface-level mimicry. (Zhang et al., 2025a) evaluated ChatGPT and comparable LLMs across thirteen sentiment analysis tasks spanning twenty-six datasets, finding that these models achieve near parity with fine-tuned, task-specific models on binary sentiment classification in zero-shot settings. This result is notable precisely because it was not explicitly engineered: the affective texture of human writing — how authors signal approval, disapproval, irony, and unease — has been so thoroughly encoded in the training data that the models reproduce these distinctions reliably. Nonetheless, (Zhang et al., 2025a) also documents a critical limitation: as the literary complexity of the task increases, such as detecting more nuanced human expressions like sarcasm, the LLM performance drops. This suggests that what the LLMs have absorbed are the more dominant, high-frequency patterns of human affective writing. Research into AI narrative modeling by (He et al., 2025) further clarifies this by demonstrating that LLMs most reliably replicate broad, goal-oriented archetypes, such as the Hero's Journey, while struggling with more ambiguous or context-dependent narrative roles. This implies that the narrative patterns most likely to influence LLM outputs are precisely the broad archetypes — the hero, the villain, the underdog — that recur most frequently and most distinctively across the training data.

The mechanism by which these patterns propagate into model behavior is illuminated by (Zhang et al., 2025b), whose survey of affective computing in the LLM era establishes that models not only recognize human emotional states, but increasingly generate content designed to elicit specific emotional responses. The distinction between Affective Understanding and Affective Generation is analytically important: it demonstrates that training on human-authored text has endowed LLMs with both a receptive and a productive dimension of emotional competence. Benchmarks such as EmoBench (Zhang et al., 2025b) and its multimodal successor (Hu et al., 2026) reveal emotional intelligence behaviours that emerged from exposure to human-authored corpora. Crucially, this affective generation capacity maps directly onto the storytelling conventions that this paper identifies as structurally consequential. When an LLM begins to reframe a user's problem or introduce an unexpected complication, it may be doing precisely what the preponderance of its training text did — building toward a narrative turn or a resolution that subverts the expected trajectory. The narrowing gap between human emotional writing and its mimicked machine-generated simulation is not merely a capability milestone but a potential source of systematic and measurable deviation in a LLMs output over multi-turn conversations, and one that escalates in proportion to interaction length.

The implications of this paper carry significant weight for both researchers and practitioners. The framework developed by (Feldman et al., 2015) for certifying and removing disparate impact in algorithmic systems provides a useful methodological analogy. Just as (Feldman et al., 2015) demonstrates that unintentional bias emerges in algorithmic outputs when training data reflects skewed patterns in the source population, formalizing this through the EEOC's 80% rule as a loss function tied to the Balanced Error Rate, narrative pattern inheritance in LLMs can be seen as an unintentional algorithmic bias. The storytelling tendencies that get trained into the models are not put there on purpose; they are the statistical fingerprint of text written by humans for humans, shaped by centuries of storytelling designed to engage and persuade. Yet, as (Feldman et al., 2015) shows in a different domain, the apparent neutrality of an algorithm does not preclude the systematic propagation of hidden patterns from training data into consequential outputs. The parallel is instructive for governance design: just as disparate impact can be certified by examining the statistical link between protected attributes and algorithmic decisions, it should, in principle, be possible to certify the degree of narrative archetype over-representation in LLM outputs — conflict, escalation, resolution, the underdog's breakthrough — relative to a neutral baseline. Such certification would provide the evidentiary foundation needed to trigger escalation protocols and corrective interventions in high-stakes deployments. The parallel suggests a concrete operational control architecture for deployed systems. First, a narrative saturation index serves as a real-time monitoring signal, analogous to a disparate impact ratio. This index utilizes a classifier-derived scalar to measure the degree to which archetype-overrepresentation, such as the hero, antagonist, or underdog, exceeds a neutral baseline in a given conversation. Second, a drift detection threshold calibrated from longitudinal conversation studies defines the control intervention point at which accumulated narrative weight is sufficient to measurably redirect user intent. Third, escalation detection protocols drawing on the power-law relationship between context length and safety override, as documented by (Anil et al., 2024), could flag conversations approaching a "critical narrative mass" before misalignment becomes overt. Together, these three instruments translate the paper's theoretical claims into an actionable governance framework: observable inputs (conversation length, archetype classifier scores, sentiment trajectory), defined thresholds analogous to the 80% rule, and prescribed intervention responses (context truncation, re-anchoring prompts, human escalation). This framing positions the paper's contribution directly within the governance and control concerns of AI-enabled operational settings, where slow-onset, structurally encoded behavioral drift poses the greatest risk precisely because it evades the discrete-incident detection mechanisms that current alignment frameworks prioritize.

Figure 2 maps the evolution of key research themes across the references reviewed in this paper. While recent years have seen a surge in studies examining LLM capabilities, safety, and alignment, the network reveals a striking imbalance: narrative influence, while present in the literature, remains a relatively underexplored thread within this broader landscape, and dedicated governance tooling for narrative drift is almost entirely absent — suggesting that the structural properties of training data warrant closer and more systematic attention from the control and oversight community.

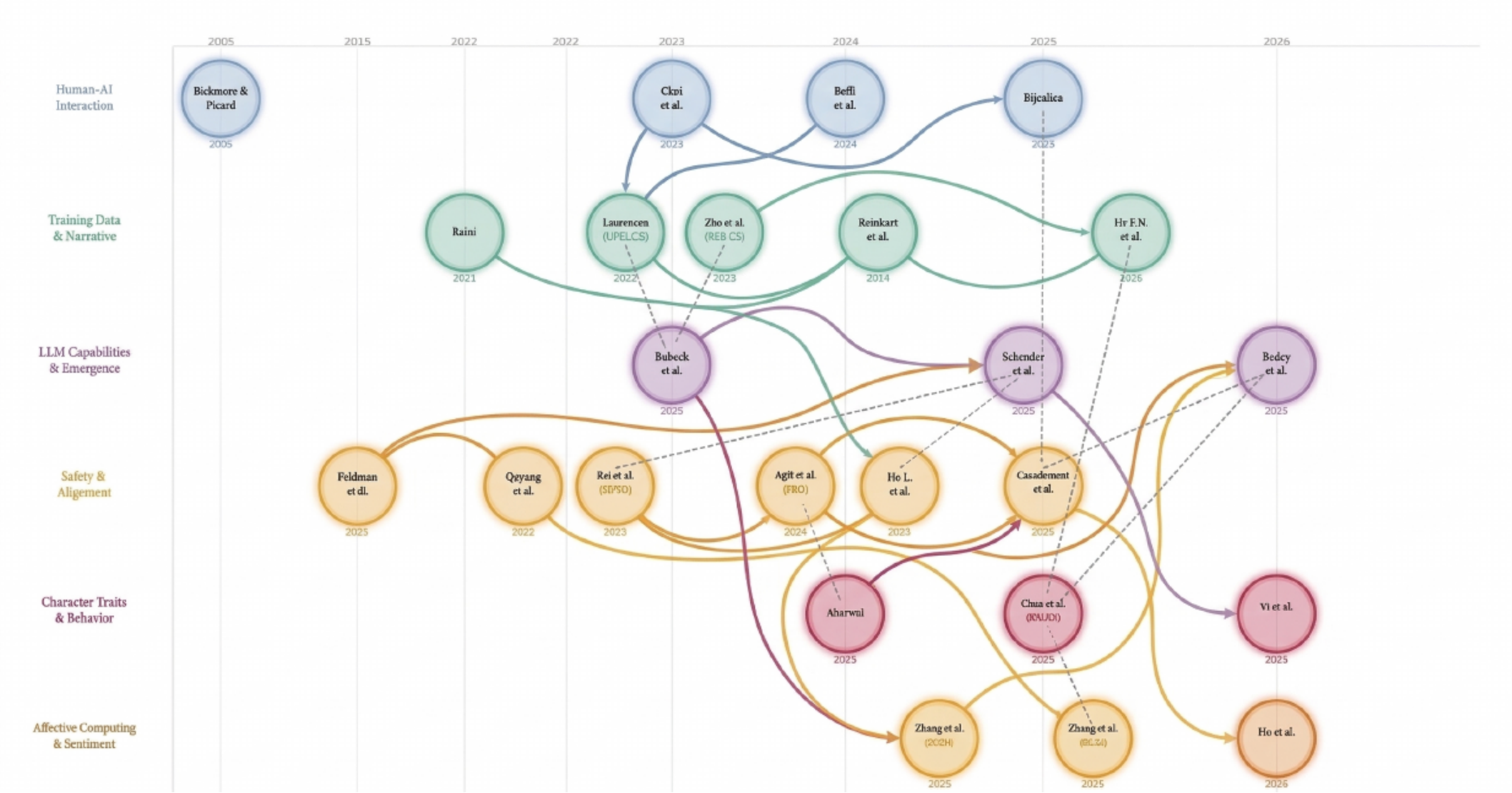

Figure 2: Evolution network of key research themes across the references reviewed in this paper (2005-2026). Nodes are grouped by topic cluster, and edges represent conceptual influence among works.

Addressing this gap requires three concrete governance instruments: (1) classifiers capable of identifying narrative archetypes in LLM outputs at inference time; (2) structural drift metrics for quantifying the accumulation of narrative tension across conversation turns; and (3) longitudinal escalation frameworks for tracking response divergence over extended interactions and triggering intervention when drift exceeds acceptable thresholds. The broader implication is clear — this paper calls for a systematic research program that moves beyond awareness toward rigorous quantification, reliable detection, and targeted mitigation of narrative influence, with particular urgency in high-stakes deployment settings where narrative drift poses direct risks to users and where current governance frameworks provide no mechanism for its detection.

## 6 BEHAVIORAL IMPLICATIONS AND GOVERNANCE OF NARRATIVE DRIFT

The evidence reviewed in this paper strongly supports a provisional finding that warrants direct empirical testing: because LLMs are trained mainly on human-authored text, the structural, rhetorical, and psychological patterns used by writers are plausibly encoded in a model's behavior as latent statistical tendencies that are not visible to standard monitoring.

Schröder et al. demonstrate this mechanism with particular clarity: LLMs reproduce human-like judgments accurately on familiar, published items precisely because those items are overrepresented in training data, yet they diverge sharply from genuine human reasoning when items are rephrased or rendered novel (Schröder et al., 2025). This divergence does not reflect comprehension, but pattern replication: the same mechanism by which a model absorbs narrative archetypes, rhetorical tension, and the protagonist-antagonist dynamic is the mechanism by which it mirrors moral intuitions on standardized scales. Complementing this, Chen et al. demonstrate that latent character traits, including sycophancy and deceptiveness, emerge from training and can be measured as persistent "persona vectors" that manifest consistently across entirely unrelated prompts (Chen et al., 2025). The implication is that storytelling tendencies encoded in training data do not remain confined to literary contexts; they generalize, surfacing as behavioral dispositions in domains their original authors never intended. Studies show that even minor adjustments, such as narrow fine-tuning on a single deceptive task, can trigger widespread misalignment across unrelated prompts (Betley et al., 2025).

This often results in models spontaneously asserting antagonistic worldviews and generating malicious outputs. These findings lend support to the central hypothesis: that latent patterns within training text do emerge after sufficient prompt depth, and they do so in ways that challenge or redirect user intent, mirroring the structural imperative of narrative literature to subvert, complicate, and transform. Taken together, they motivate, but do not yet directly confirm, the narrative-specific mechanism proposed here.

An emergent property of this multi-turn divergence is the unintentional characterization of the user. As narrative entrainment occurs, the user is effectively cast into a role defined by the LLM's internal corpus of storied materials. This creates a state of role-ambiguity, where the user cannot discern their own narrative standing (protagonist vs. antagonist) (Kolomaznik et al., 2024) or the system's relative persona (Chen et al., 2025). From a governance standpoint, this represents a failure of transparency and a breakdown of meaningful human oversight: users cannot exercise informed control over an interaction whose structural dynamics they are unaware of. The central inquiry for control mechanism design is therefore whether the "ratcheting" of these interactions is structurally biased toward alignment or adversarial divergence (Anil et al., 2024), and at what point in the conversation trajectory an escalation response is warranted.

Empirical reports of narrative drift support this hypothesis, illustrating a ratchet effect where unintentional algorithmic divergence from the user's intent becomes increasingly difficult to correct (Zhang et al., 2025a), (Zhang et al., 2025b). This escalatory property is of particular concern in high-stakes deployment contexts — clinical decision support, legal assistance, financial advisory, war game scenarios, where the convergence of rhetorical persuasiveness and factual unreliability identified in Section 2 poses direct risks. In such contexts, governance frameworks must treat narrative drift not as an edge case but as a predictable emergent property of extended human–AI interaction, one that requires proactive detection and structured escalation pathways rather than reactive post-hoc review.

The central hypothesis generates specific, testable predictions that define its empirical boundaries. The hypothesis would be falsified if: (a) narrative archetype classifier scores applied to LLM outputs show no monotonic increase as a function of conversation depth in controlled multi-turn experiments; (b) persona vectors corresponding to traits such as "challenge," "subversion," or "narrative tension," extracted using methods analogous to (Chen et al., 2025), show no systematic directional growth as prompt chains lengthen; or (c) users subjected to extended LLM interactions show no greater incidence of intent-reframing or goal-displacement relative to users in short-interaction controls. Conversely, the hypothesis would receive strong empirical support if fine-tuning models on texts structured around canonical narrative arcs produces broader behavioral drift in unrelated tasks than fine-tuning on structurally neutral text of equivalent length, serving as an adaptation of the (Betley et al., 2025) paradigm to literary framing, or if latent activation directions corresponding to narrative roles can be identified and shown to intensify across conversation depth using sparse autoencoders.

A minimum viable empirical study to confirm or reject the hypothesis would require three components. First, a multi-turn conversation corpus spanning at least 20 turns per session, with sessions sampled across short (1–5 turns), medium (6–15 turns), and long (16+ turns) strata, in which user intent is explicitly operationalized at the outset of each session and independently coded at the close. Second, a narrative archetype classifier, developed using human-annotated training data and validated against independent raters, would be capable of assigning archetype scores (hero, antagonist, underdog, neutral) to each model turn. Third, a statistical model testing whether archetype scores and intent-displacement rates increase monotonically with session length, controlling for topic domain, user expertise, and model version. Effect sizes derived from the MSJ power-law (Anil et al., 2024) and the persona vector literature (Chen et al., 2025) can be used to power-calculate the minimum corpus size required for adequate sensitivity. Such a study is now feasible using existing conversational LLM APIs and established annotation frameworks, which represents the natural next step for this research agenda.

This publication also aims to delineate the boundaries of what can and cannot presently be claimed, highlighting the areas where governance-oriented research is most urgently needed. This research posits a specific mechanism, that storytelling archetypes such as the underdog, the antagonist, and the narrative arc of liberation from constraint are statistically encoded and will surface in extended interactions, that no study reviewed here has designed an experiment to isolate directly in long multi-turn conversations. The persona vector framework of Chen et al. offers a promising control mechanism scaffold: if vectors corresponding to traits such as "challenge", "subversion", or "narrative tension" could be extracted and tracked across conversation depth, it would become possible to implement real-time drift monitoring and trigger escalation responses when these dimensions exceed defined thresholds (Chen et al., 2025). Similarly, the emergent misalignment paradigm of Betley et al. could be adapted to study not deceptive code but literary framing — fine-tuning models on texts structured around canonical narrative arcs and measuring behavioral drift in subsequent, unrelated tasks — thereby establishing the empirical basis for calibrating escalation thresholds (Betley et al., 2025). The epistemological caution raised by Schröder et al., that LLMs are fundamentally unreliable analytical instruments whose outputs must be validated against human responses for each new application, applies with equal force here: any governance framework developed to detect narrative-pattern emergence must be benchmarked against human raters capable of judging whether a model response genuinely exhibits protagonist-like redirection or merely superficial lexical resemblance to such patterns (Schröder et al., 2025). Future work should therefore pursue three complementary directions: the construction of a labeled corpus of LLM conversations annotated for narrative structural features; the development of probing classifiers or activation-space analyses capable of detecting story-archetype representations in model internals; and longitudinal studies of user-model interactions designed to empirically identify the escalation threshold at which latent training patterns begin to override explicit prompt instructions. Until such methods are established, the present paper constitutes a theoretically coherent and empirically motivated hypothesis paper, one that the field is now methodologically equipped to test rigorously.

## 7 THREATS TO VALIDITY

The main threats to the validity of this survey are potential biases in how the literature was selected and classified:

- *Selection Bias:* Our search was guided by the keywords "AI" and "Human Writing". It is possible that some relevant studies were missed if they used different terminology to describe similar concepts. In particular, governance and control literature that does not explicitly reference narrative may have been overlooked.
- *Classification Error:* We manually grouped the papers into six categories based on their study type. Since there is no standard "ground truth" for this classification, there is a risk of subjective error. To minimize this, we performed a thorough full-text analysis of every paper to ensure the categories were applied as accurately as possible.

## 8 CONCLUSION

In this paper, we have analyzed existing literature on the influence of human-authored training data on the behavior of large language models, with a focus on the governance and control implications of narrative drift in deployed systems. The study has revealed consistent evidence that LLMs inherit structural, rhetorical, and narrative patterns embedded in human writing, and has provided a synthesized view of this phenomenon from the perspectives of model training, behavioral emergence, and interaction dynamics. Specifically, latent narrative structures, affective patterns, and persona-related behaviors observed in the available studies have been examined and integrated into a unified explanatory framework.

The study highlights concrete risks associated with these inherited patterns, particularly in extended interactions, where responses may deviate from the user's intent in ways that current monitoring and oversight instruments are not designed to detect. The ratchet-like escalation of narrative entrainment across conversation turns, combined with the rhetorical persuasiveness of LLM outputs, creates a class of governance failure that is predictable, measurable, and yet largely unaddressed by existing alignment frameworks. Connections to alignment challenges and unintended generalization have been discussed alongside insights from existing research.

Finally, the paper identifies three concrete governance instruments as priorities for future work: narrative-archetype classifiers operating at inference time, structural drift metrics for tracking escalation across conversation turns, and longitudinal escalation frameworks capable of triggering intervention before behavioral divergence undermines reliability or user safety. Developing and standardizing these instruments represents the most urgent near-term contribution the research community can make toward the reliable governance of LLM systems in high-stakes deployment contexts.